# Machine learning for the prediction of safe and biologically active organophosphorus molecules

Hang Hu [†,*], Hsu Kiang Ooi [†], Mohammad Sajjad Ghaemi [†], Anguang Hu [‡]

[†] National Research Council of Canada (NRC)

[‡] Suffield Research Centre, Defence Research and Development Canada

Drug discovery is a complex process with a large molecular space to be considered. By constraining the search space, the fragment based drug design is an approach that can effectively sample the chemical space of interest. Here we propose a framework of Recurrent Neural Networks (RNN) with an attention model to sample the chemical space of organophosphorus molecules using the fragment-based approach. The framework is trained with a ZINC dataset that is screened for high druglikeness scores. The goal is to predict molecules with similar biological action modes as organophosphorus pesticides or chemical warfare agents yet less toxic to humans. The generated molecules contain a starting fragment of $PO_2F$ but have a bulky hydrocarbon side chain limiting its binding effectiveness to the targeted protein.

**1. Introduction**

Modern agriculture often uses fast-acting organophosphorus (OP) pesticides to help maintain plant growth, control diseases, protect from insects and other harmful organisms, and guarantee consistent plant yield[1]. When consumed, the residual pesticides in agricultural products can cause indelible damage to human health[1]. Many of these OP compounds are also structurally similar to many chemical warfare agents (CWAs), such as sarin. There is a growing concern over malicious individuals gaining access to weaponized OP agents with the intent to cause mass civilian casualties. The victim of CWAs in the recent Syrian civil war and the Rhodesian war, as well as the death of Indian children following the consumption of pesticide-contaminated lunches, are just some of the recent incidents of public exposure to these toxic chemicals[1–4].

In general, weaponized OP compounds often target the nervous system by inhibiting acetylcholinesterase (AChE) and phospholipase, which leads to a build-up of neurotransmitters at the nerve synapses[1]. This is the leading cause of the high toxicity of OP-based CWAs[5]. As a result, the victim will initially experience cholinergic symptoms, such as salivation, lacrimation and seizures, ultimately resulting in rapid death if untreated. In addition to regulating the production and excessive application of these toxic chemicals, which can damage the environment on top of adverse effects on human health, efficient detection methods are essential for food and public safety monitoring systems[6, 7]. Among the many OP compounds, paraoxon is commonly found as a paradigm and substitute for CWAs[8–10], but paraoxon's toxic nature and uncertain optical properties left much to be desired. Therefore, it is crucial to develop safe, protein-inhibiting OP compounds with a similar mode of action that can be substituted for CWA for research laboratories and regulatory agencies[11–13].

*hang.hu@nrc-cnrc.gc.ca

## 2. Relevant Prior Works

Over the past decade, machine learning-based de novo molecule designs have gained increased popularity and have been applied in various systems[14,15]. One common approach is to adopt natural language processing methods, such as recurrent neural networks (RNN) with self-attention have shown the ability to overcome typical energy space barriers and predict novel molecules with good chemical sense. Part of RNN's popularity can be attributed to the LSTM (long short-term memory) developed in 1997[15,16]. LSTM cell helps RNN consider the input sequence's long- and short-term dependency. Furthermore, the now famous "attention" mechanism was first introduced in 2014[17,18]. The attention layer creates a context vector based on the calculated attention weights. This helps the model decide which of the previous states is more impactful for the next prediction. Our approach utilizes an attention layer and LSTM to prepare model training and combined with fragment-based molecule generation for targeted drug discovery, yielding molecules with similar behaviour as known ligands but with more desirable properties. Compared to the popular de novo molecular method, our approach alleviates the need for the time-consuming model fine-tuning procedures and is suitable for the initial stage of drug discovery[17].

## 3. Proposed methods

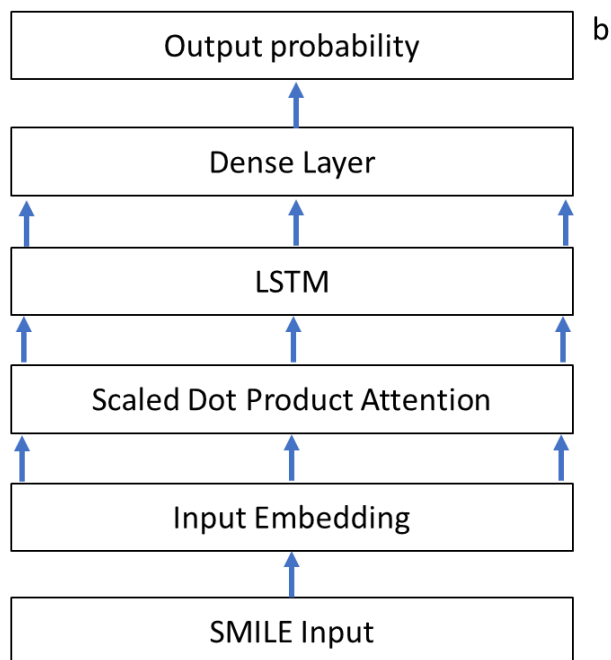

Figure 1. (a) Schematic of the training procedure of the RNN model. The model will read the input smile sequence and attempt to predict the next entry. (b) Flowchart of the RNN-self attention framework.

We screened the Zinc dataset and obtained 0.5 million molecules with a quantitative estimate of the druglikeness (QED) score (>0.65). These molecules were converted into SMILE strings, which were then tokenized for RNN model training. The loss function is calculated from the accuracy of the prediction. The RNN model contains an embedding layer, an attention layer, an LSTM layer and followed by a dense layer. The attention layer is designed with a softmax activation function and a hyperbolic tangent activation function for the attention weight and bias. At the start of training, the attention weight is initialized with normal distribution, and the bias is all 0. To train the model, we break each input into an overlapping sequence of tokens with a fixed length, and the RNN model predicts the next token of the sequence, as illustrated in Figure 2. After the model had demonstrated the ability to predict valid SMILE strings, we adopted a fragment-based growing process. The starting fragment is guaranteed to contain "P", "F", "O" and "C". The model "grows" the molecule based on the tokenized starting string. We generated 1200 molecules of various lengths and calculated their properties, such as octanol-water partition coefficient (log P), intrinsic hepatic clearance, Caco-2 cell permeability and QED score. The properties are calculated using the QSAR model implemented under the OPERA package[19]. We also studied the likelihood of the generated molecules binding to biological phospholipase. To this end, we performed rigid docking of the molecules to bovine brain platelet phospholipase A2(3DT8) using Autodock Vina[20, 21].

## 4. Results and Discussion

The quantitative estimate of the druglikeness score (QED) was the main factor used to construct the training set. From the predicted molecules, we also observed clustering of QED score demonstrating the generational improvement associated with RNN with attention layer, as shown in Figure 2. The first generation of the predicted molecules has an average QED score of 0.37, the second generation has an average QED score of 0.55, and for the third generation, the average QED score improved to 0.66. We observed no significant changes to the average QED score beyond the third generation. The molecular structure of sarin and seven other derived molecules with the highest QED score are shown in Figure 3.

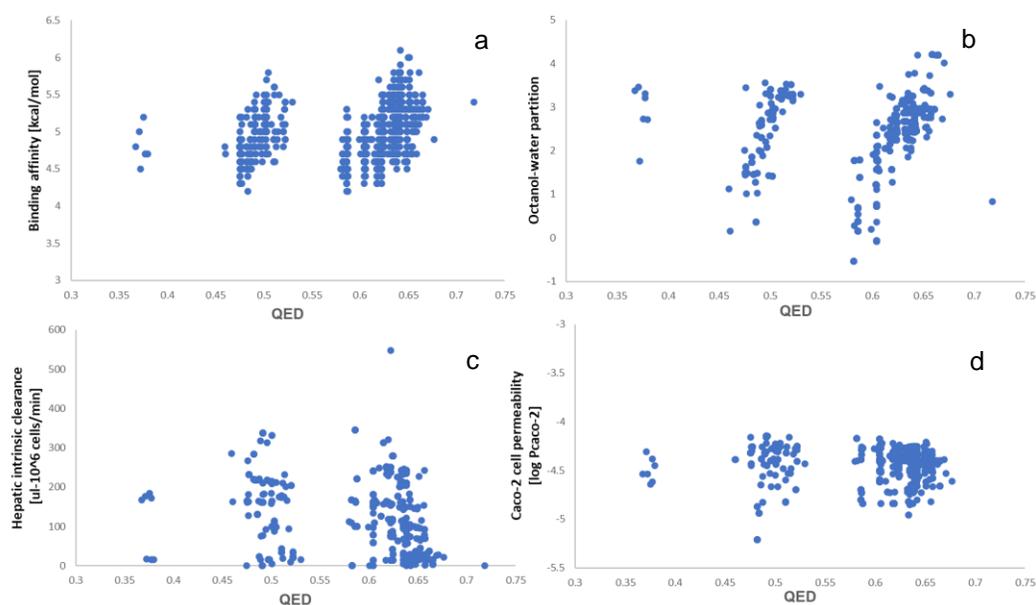

*Figure 2. The plot of various molecule properties versus QED. For (a) the binding affinity is calculated from rigid docking with Autodock Vina, (b) octanol-water partition coefficient, (c) intrinsic hepatic clearance and Caco-2 cell permeability are calculated using the QSAR model implemented under the IntrinsicOPERA package.*

For the calculated binding affinities to phospholipase, the docking score for all the molecules varies from 4 to 6 kcal/mol. The generated molecules are grown from a key $PO_2F$ functional group of sarin, shown in Figure 3. The binding affinity of sarin to the phospholipase is 12 kcal/mol. The binding affinity suggests that most molecules can form weaker interactions with the active site of the phospholipase protein. The low binding affinity is essential for this research project, as the goal is to design molecules with similar biological functions as organophosphorus-based CWAs with less potency. Compared to sarin's molecular structure, we have intentionally designed the molecule generation process to increase the size of the hydrocarbon chain connecting the $PO_2F$ functional group. This bulky chain will likely hinder the molecules from binding to the active site of the phospholipase.

The QED gives a general likelihood of a particular molecule behaving like drugs in physiological conditions. In addition to QED, we also looked at the specific molecule characteristics, octanol-water partition coefficient, intrinsic hepatic clearance and Caco-2 cell permeability. All the generated molecules have an octanol-water partition coefficient of less than 5, indicating they are likely to accumulate in the fatty tissues[22]. The intrinsic hepatic clearance measures the rate a liver removes a drug from circulation inside a body, and Caco-2 cell permeability measures a drug's gastrointestinal (GI) absorption. For intrinsic hepatic clearance rate, a typical range for most of the drugs is above 300 ul-106 cells/min[23]. Most generated molecules have intrinsic hepatic clearance rates below 300 ul-106 cells/min. This is expected as long single-bonded hydrocarbon chains with bulky rings can be more complicated to break down, as shown in Figure 3. Typically, a log scale Caco-2 cell permeability greater than six suggests that 50% to 100% of the drugs are likely to be absorbed by the GI tract [24], and the Caco-2 cell permeability of all the generated drugs is less than 6. For our purposes, this is a potential advantage as the danger of accidental oral ingestion is diminished.

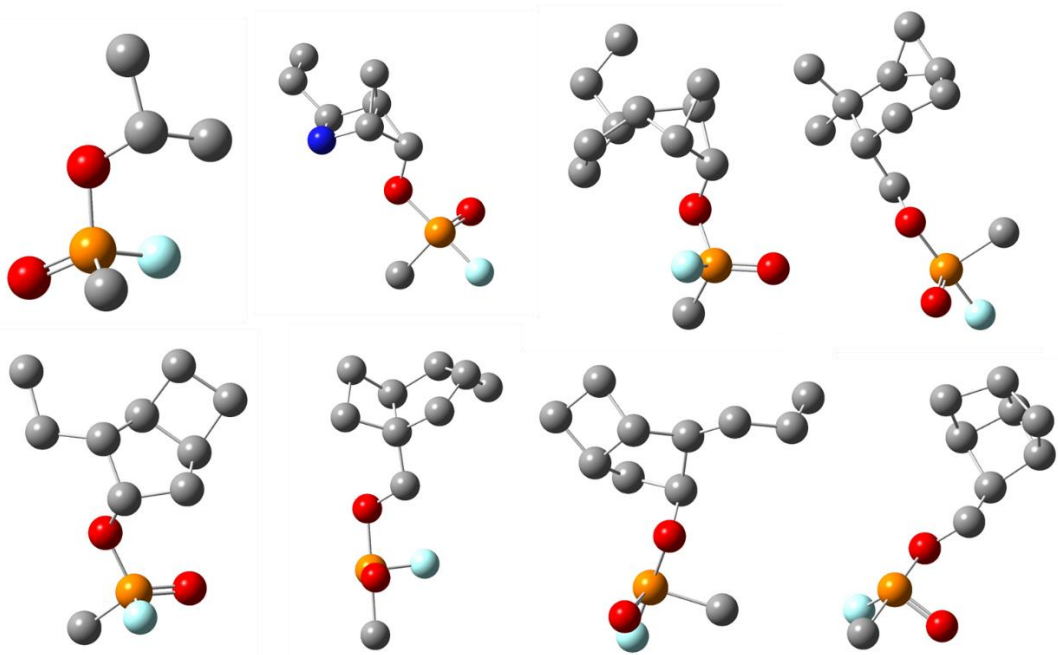

*Figure 3. Molecular structure of sarin (top left) and seven other generated molecules. Grey spheres represent carbon atoms, red spheres represent oxygens, light green spheres represent fluorine, blue spheres represent nitrogen, and yellow spheres represent phosphorus.*

## 5. Conclusions

The RNN with attention model with fragment-based molecule generation is an efficient approach for *de novo* molecule design with targeted properties. The machine learning model can "learn" patterns that generate valid SMILE string based molecular structures. The general design of the target molecule can be largely biased with the initial starting fragments. By applying domain knowledge (QED score) when developing the training set, we successfully obtain an average QED score of 0.65. The distinct cluster of the QED score of the molecules shows generational improvement. The preferred OP molecules with high QED scores often have bulky hydrocarbon rings connecting to the $PO_2F$ functional group. These resulting molecules can still interact with phospholipase protein, but the binding affinity is significantly reduced. The generated molecules can bioaccumulate and have a low potential GI absorption rate. Future work with more fine-tuning of the training procedure could improve the average QED score of the generated molecules. Domain knowledge, such as ease of synthesis, could be added to the training set to improve the chemical intuition of the training model.

## Acknowledgements

This project is supported by the National Research Council Canada (NRC) under the AI for Design Challenge Program and the Defence Research and Development Canada (DRDC).